\documentclass[cameraready]{Interspeech}

\usepackage{multirow}
\usepackage{soul, xcolor}

\usepackage{caption}
\usepackage{subcaption}
\usepackage{graphicx}

\title{Hybrid Continual Learning for Low-Resource Australian Aboriginal Language Identification}

\author[affiliation={1}, orcid=0009-0005-3439-4970]{Pravina}{Mylvaganam}
\author[affiliation={2}]{Ting}{Dang}
\author[affiliation={1}]{Eliathamby}{Ambikairajah}
\author[affiliation={1}]{Vidhyasaharan}{Sethu}
\author[affiliation={3}]{Jingyao}{Wu}

\address{
    $^1$ University of New South Wales, Australia \\
    $^2$ University of Melbourne, Australia \\
    $^3$ Massachusetts Institute of Technology, USA 
}

\email{p.mylvaganam@unsw.edu.au , ting.dang@unimelb.edu.au, e.ambikairajah@unsw.edu.au, v.sethu@unsw.edu.au, jingyaow@mit.edu}


\usepackage{comment}

\begin{document}

\maketitle

\begin{abstract}
\vspace{-5pt}

Language identification is an important step toward integrating endangered Australian Aboriginal languages (AALs) into speech technologies supporting language revitalisation and digital inclusion. However, extreme data scarcity limits model performance. Transfer learning from high-resource languages shows promise but often suffers from catastrophic forgetting when adapting to new languages. Continual learning (CL) can mitigate this issue, though it remains challenging with very limited data. To address this, we propose two hybrid continual learning methods: Replay Augmented Elastic Weight Consolidation and Constraint Guided Knowledge Distillation to adapt pretrained speech models for AAL identification while preserving previously learned knowledge. Experiments on Warlpiri, Dalabon and Dharawal show that the proposed methods outperform fine-tuning and existing CL baselines, improving adaptation to multiple AALs while maintaining performance on previously learnt high-resource languages.

\end{abstract}

\section{Introduction}
\vspace{-5pt}

Australia has over 250 Indigenous languages, but only about 123 remain actively used, many critically endangered due to limited intergenerational transmission \cite{battin2020national}. As language loss risk grows, technological support is vital for documentation and revitalization. Language identification (LID), which automatically detects the language of an audio signal, plays a key role in this effort. It enables downstream applications, such as speech recognition, translation, and speech archiving, while supporting efficient corpus organization. Reliable LID is thus crucial for developing digital tools that promote preservation, education, and inclusion for Australian Aboriginal communities.

Recent advances in multilingual speech modeling, supported by large-scale speech corpora, have enabled near-perfect LID for high-resource languages \cite{valk2021voxlingua107}. However, these gains have not transferred effectively to low-resource settings, where performance remains substantially lower due to severe data scarcity \cite{caswell2020language}. The challenges are even more pronounced for Australian Aboriginal languages (AALs), which often exhibit restricted speaker diversity and limited access to linguistic expertise.

Prior work on low-resource languages has explored domain adaptation via full or parameter-efficient fine-tuning \cite{feng2019low, gaikwad2021cross, hu2022lora}. However, these methods often cause catastrophic forgetting, where performance on previously learned languages drops significantly when adapting to new ones. This issue is critical in multilingual systems that must support both high-resource languages and newly introduced AALs. Moreover, repeatedly fine-tuning for each language is computationally costly, emphasizing the need for methods that allow continuous adaptation while retaining performance on previously learned languages.


Continual learning (CL) addresses these challenges by enabling models to adapt to new data while retaining previously learned tasks. CL methods are typically categorized into three groups \cite{qu2021recent}: (i) regularization-based, such as Elastic Weight Consolidation (EWC) \cite{EWC}, which constrains parameter updates to preserve previously learned tasks; (ii) replay-based, like Experience Replay (ER) \cite{fedus2020revisiting}, which rehearse representative samples from earlier tasks; and (iii) knowledge-distillation (KD) methods, which transfer knowledge from a previous model to mitigate forgetting \cite{mansourian2025comprehensive}. However, applying these methods to AALs remains challenging. Regularization approaches become less effective during long-term adaptation to diverse AALs, as variations in accents, acoustic conditions, and languages can interfere with prior knowledge \cite{van2019three}, and accumulated penalties over time may cause numerical instability \cite{jones2018continual}. ER and KD also struggle with data imbalance between newly introduced and previously learned languages, common in extremely low-resource AALs, which can hinder adaptation or destabilize prior knowledge \cite{qu2021recent, hou2019learning}, indicating that ER or KD alone is insufficient to prevent catastrophic forgetting in such settings.


Recent studies have explored hybrid approaches that combine multiple CL methods to overcome the limitations of individual ones in other domains \cite{li2024continual}. For example, Sun et al. \cite{sun2020distill} and Cappellazzo et al. \cite{cappellazzo2022investigation} propose that combining ER and KD improves accuracy and reduces forgetting compared to standalone methods, particularly in class-incremental settings for natural language and spoken language understanding. However, hybrid strategies remain largely unexplored for LID, especially for highly under-resourced languages. Existing ER–KD hybrids may also be well-suited for AALs, as they remain sensitive to data imbalance \cite{qu2021recent, hou2019learning}, a key challenge in AALs. Consequently, such methods may still suffer from biased learning and forgetting, underscoring the need for hybrid CL frameworks specifically designed for extreme low-resource settings.

To address limitations for extremely low-resource AALs, we propose two hybrid continual learning frameworks to overcome severe data scarcity and catastrophic forgetting: (1) Replay-Augmented Elastic Weight Consolidation (RA-EWC), which augments EWC, which preserves previously acquired knowledge by restricting changes to important model parameters, with selective replay of past samples to reinforce earlier learning, and (2) Constraint-Guided Knowledge Distillation (CG-KD), which integrates EWC with KD, where a previously trained model guides the learning of a new model through soft target predictions, for robust adaptation under extreme scarcity. These methods are especially well-suited for AALs, as RA-EWC supports stable learning even when replay memory is very limited, while CG-KD protects important model parameters and preserves the behavior of previously trained models, enabling more reliable adaptation to new languages.

We evaluate the proposed methods on three extremely low-resource AALs: Warlpiri, Dalabon, and Dharawal, and on two different tasks: (i) single AAL adaptation, which extends a pretrained model to learn a single AAL, and (ii) sequential adaptation across multiple AALs, which enables incremental learning without forgetting. To our knowledge, this is the first work to introduce hybrid CL methods for extremely low-resource language identification tasks.

\section{Proposed Hybrid Continual Learning}
\vspace{-5pt}

\begin{figure}[t]
    \centering
    \includegraphics[width=0.7\linewidth]{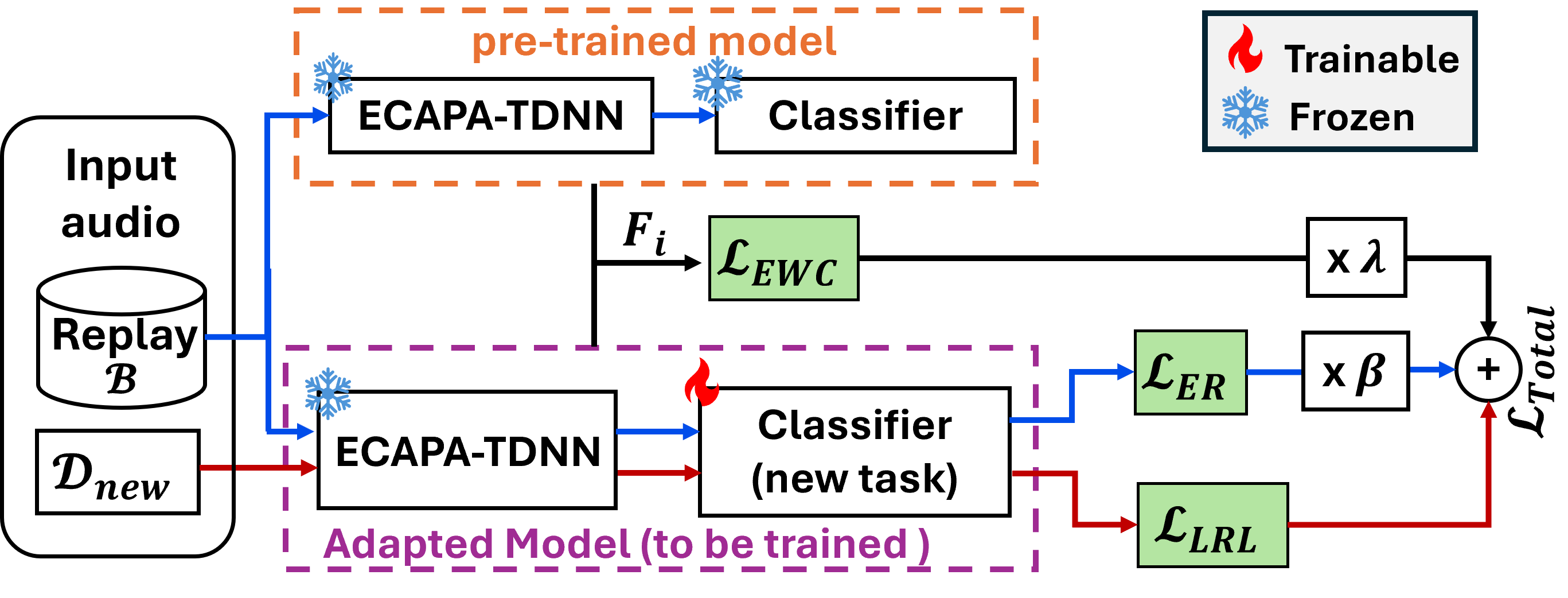}
    \vspace{-4mm}
    \caption{Overview of Replay-Augmented Elastic Weight Consolidation (RA-EWC), combining EWC and ER. It uses three losses: negative log-likelihood $\mathcal{L}_{\text{LRL}}$ on new AAL data, replay loss $\mathcal{L}_{\text{ER}}$ on stored replay samples, and regularization constraint $\mathcal{L}_{\text{EWC}}$ to protect important parameters. All losses are jointly optimized to adapt the model.
    }
    \label{fig:RA-EWC}
    \vspace{-15pt}
\end{figure}

We propose two hybrid methods, RA-EWC and CG-KD, to address the challenges of low-resourced AAL identification. RA-EWC enhances EWC by incorporating ER, as illustrated in Figure~\ref{fig:RA-EWC}. This hybrid design allows the model to retain important parameters learned from prior languages ($F_i$) even with a limited replay sample, stabilizing learning and mitigating catastrophic forgetting during continual adaptation of AALs. As shown in Figure~\ref{fig:2}, the proposed CG-KD combines KD with EWC to guide continual learning. The student model learns representations for the target low-resource languages while preserving its outputs on high-resource languages to match those of a frozen teacher model (i.e., the pretrained model prior to continual learning). 


\subsection{Replay-Augmented Elastic Weight Consolidation  (RA-EWC)}
\vspace{-5pt}


\noindent \textbf{Constraint on the representative buffer. } By maintaining a small replay buffer $\mathcal{B}$ containing samples from previously seen high-resource languages $\mathcal{D}_{\mathrm{HRL}}$, forgetting can be mitigated by periodically re-exposing the model to the source domain~\cite{fedus2020revisiting}. During continual adaptation to AALs, each training batch comprises both newly observed low-resource data $\mathcal{D}_{\mathrm{LRL}}$ and replayed samples $\mathcal{B} \subset \mathcal{D}_{\mathrm{HRL}}$. Accordingly, the adaptation objective combines the negative log-likelihood (NLL) loss for low-resource data, $\mathcal{L}_{\mathrm{LRL}}$, with the NLL loss for high-resource replay data, $\mathcal{L}_{\mathrm{ER}}$, as follows:

\vspace{-15pt}
\begin{align}\label{eq:2}
\mathcal{L}_{\text{LRL}}(\theta) &= \mathbb{E}_{(x,y)\sim\mathcal{D_\text{LRL}}}\left[\mathcal{L}\big(f_\theta(x),y\big)\right]
\end{align}

\vspace{-21pt}
\begin{align}\label{eq:1}
\mathcal{L}_{\text{ER}}(\theta) &= \mathbb{E}_{(x,y)\sim\mathcal{B}}\left[\mathcal{L}\big(f_\theta(x),y\big)\right]
\end{align}

\noindent Here, $f_\theta(x)$ denotes the output of continually adapted model for given input $x$ and $y$ represent the corresponding groundtruth.

\noindent \textbf{Constraint on critical weights. } Penalizing changes to parameters that are important for previously learned high-resource languages, $\mathcal{D_\text{HRL}}$, can help constrain weight updates and prevent catastrophic forgetting. The importance of each pretrained parameter, $\theta_i$, is quantified using an importance score $F_i$, which is the $i$-th diagonal element of the Fisher Information Matrix, $\mathbf{F}$ \cite{kirkpatrick2017overcoming}. The $\mathbf{F}$ matrix assigns higher values to parameters that strongly influence the loss, indicating that changes to these parameters significantly affect performance on previously learned tasks, while parameters with lower values are less critical. Let $\theta_i^*$ denote the updated model parameters during continual learning, and therefore, the regularization loss is:

\vspace{-5mm}
\begin{align}\label{eq:3}
\mathcal{L}_{\text{EWC}}(\theta) &= \sum_i F_i (\theta_i^* - \theta_i)^2
\end{align}
\vspace{-12pt}

Minimizing the loss function ensures that important parameters remain close to their previous values, reducing forgetting while continuously learning new languages.

\noindent \textbf{Total loss. }The total loss is the combination of the NLL loss for new language, $\mathcal{L}_{\text{LRL}}$ and the regularization losses as: 

\vspace{-15pt}
\begin{align}
    \mathcal{L}_{Total} (\theta) &= \mathcal{L}_{\text{LRL}}(\theta) + \beta \, \mathcal{L}_{\text{ER}}(\theta) + \lambda \, \mathcal{L}_{\text{EWC}} (\theta)
\end{align}
\vspace{-15pt}

\noindent where $\lambda$ and $\beta$ are hyperparameters balancing the trade-off between the terms.

\subsection{Constraint-Guided Knowledge Distillation (CG-KD)}
\vspace{-5pt}
\begin{figure}[!t]
    \centering
    \includegraphics[width=0.7\linewidth]{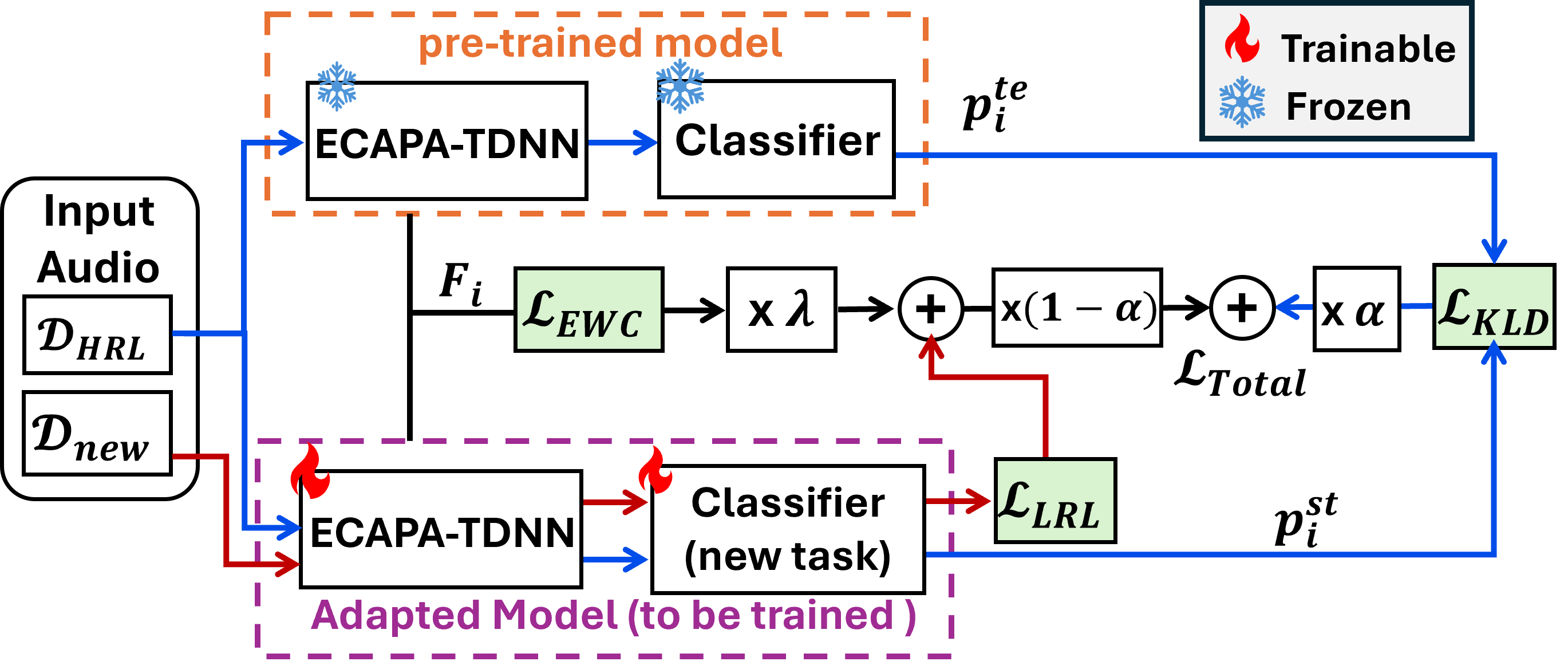}
    \vspace{-4mm}
    \caption{Overall architecture of Constraint-Guided Knowledge Distillation (CG-KD), integrating KD and EWC. The student learns from new AAL data and aligns its high-resource outputs with the frozen teacher via $\mathcal{L}_{\text{KLD}}$ loss. The regularization term $\mathcal{L}_{\text{EWC}}$ preserves key weights, and $\mathcal{L}_{\text{LRL}}$ supports accurate classification of new languages. All losses are jointly optimized to update the model.
    }
    \vspace{-15pt}
    \label{fig:2}
\end{figure}




The Kullback–Leibler Divergence (KLD) loss is used to adapt the student model to preserve knowledge from the pre-trained teacher by aligning their output distributions over previously learned high-resource languages classes $i$:

\vspace{-15pt}
\begin{align}
\small
\mathcal{L}_{\text{KLD}}
&= - \sum_i p_i^{\text{te}}(T), \log p_i^{\text{st}}(T)
\end{align}
\vspace{-10pt}

Here, $p_i^{\text{te}}(T)$ and $p_i^{\text{st}}(T)$ denote the softened probability distributions of the teacher and student models, respectively, computed using a temperature-scaled softmax:
\vspace{-5pt} 
{\small
\begin{align}
p_i^{\text{te}}(T) &= \frac{\exp(z_i^{\text{te}} / T)}{\sum_j \exp(z_j^{\text{te}} / T)}, \quad
p_i^{\text{st}}(T) = \frac{\exp(z_i^{\text{st}} / T)}{\sum_j \exp(z_j^{\text{st}} / T)}.
\end{align}
}
\vspace{-5pt}

\noindent where $z_i^{\text{te}}$ and $z_i^{\text{st}}$ denote the teacher and student logits for high-resource language $i$, and $T$ is a temperature hyperparameter. Dividing logits by $T$ before softmax, a higher $T$ softens the output distribution, increasing the relative probabilities of non-target classes and thereby exposing similarity patterns among classes, which helps capture inter-class relationships during knowledge transfer. 

In addition, EWC is applied via the regularization loss in Equation \eqref{eq:3}, which constrains important parameters of the student model and promotes stability during adaptation. To effectively learn low-resource languages, the NLL loss $\mathcal{L}_{\text{LRL}}$ is also included to optimize classification performance on the newly introduced languages.


\noindent \textbf{Total loss. }The total loss of CG-KD is a weighted sum of the $\mathcal{L}_{\text{LRL}}$, regularization loss and the KLD loss as:

\vspace{-15pt}
\begin{equation}
    \mathcal{L}_{\text{total}}(\theta) = (1 - \alpha) \,(\mathcal{L}_{\text{LRL}}(\theta) + \lambda \, \mathcal{L}_{\text{EWC}}(\theta)) + \alpha \, \mathcal{L}_{\text{KLD}}(\theta)  
\end{equation}
\vspace{-15pt}


\noindent where $\alpha$ controls the balance between learning the new language and retaining knowledge from previously seen languages.

\begin{table*}[ht]
\footnotesize
\centering
\caption{Performance (F1-score \%) for high-resource languages (HRL) and three AALs: Warlpiri, Dalabon, and Dharawal. "Source" and "TL" refer to pre-trained model without adaptation and naïve transfer learning, respectively. Best results are highlighted in \textbf{Bold}.}
\vspace{-2mm}
\begin{tabular}{lccc|ccc|ccc}
\hline
\multirow{2}{*}{\textbf{Method}} & \multicolumn{3}{c|}{\textbf{Warlpiri (in \%)}} & \multicolumn{3}{c|}{\textbf{Dalabon (in \%)}} & \multicolumn{3}{c}{\textbf{Dharawal (in \%)}} \\
\cline{2-10}
 & HRL & Warlpiri & Overall & HRL & Dalabon & Overall & HRL & Dharawal & Overall \\
\hline
Source & 90.72 & - & - & 90.72 & - & - & 90.72 & - & - \\
TL     & 62.47 & 98.70 & 65.38 & 46.85 & 100 & 47.99 & 31.25 & 100 & 32.64 \\
\hline
\textit{CL baselines} & & & & & & & & & \\
EWC    & 84.06 & 100 & 85.36 & 52.54 & 82.35 & 54.46 & 46.02 & 100 & 50.38 \\
ER     & 84.03 & 98.70 & 85.26 & 52.50 & 82.35 & 54.28 & 46.01 & 100 & 50.30 \\
KD     & 91.56 & 100 & 93.02 & 84.12 & 100 & 84.60 & 74.96 & 100 & 75.99 \\
\hline
\textit{proposed methods} & & & & & & & & & \\
RA-EWC & 86.76 & 100 & 87.41 & 75.36 & 100 & 76.58 & 66.38 & 100 & 68.96 \\
CG-KD  & 92.89 & 100 & \textbf{93.38} & 85.33 & 100 & \textbf{85.68} & 76.47 & 100 & \textbf{76.41} \\
\hline
\end{tabular}
\vspace{-10pt}
\label{tab:combined}
\end{table*}

\section{Experimental Setup}
\subsection{Dataset}
\label{dataset}
\vspace{-5pt}
\noindent\textbf{AAL dataset. }Experiments are conducted on three AALs: Warlpiri, Dalabon, and Dharawal. Speech data for Warlpiri \cite{doreco-warl1254} and Dalabon \cite{doreco-ngal1292} is sourced from the DoReCo dataset \cite{doreco}, with recordings from 18 and 4 speakers, respectively. Dharawal recordings are obtained from publicly available resources on the Dharawal Words website\footnote{\url{https://www.dharawalwords.com.au}}\cite{dang2025characterization}, consisting of 8 speakers. Although small compared to high-resource languages, these are the largest available AAL speech datasets, highlighting severe data scarcity. All recordings were manually inspected, and files with unclear speech, silence, excessive noise, or corruption were removed to preserve linguistic integrity and meaningful embeddings. After preprocessing, audio was downsampled to 16 kHz and segmented into 8–10 second clips, yielding 1125 utterances (3 hours) for Warlpiri, 284 utterances (40 minutes) for Dalabon, and 63 utterances (11 minutes) for Dharawal. Each dataset was split 80/10/10 for training, validation, and testing.

\noindent \textbf{High-resource languages. } 
Speech data for high-resource languages is drawn from the VoxLingua107 \cite{valk2021voxlingua107}, selecting 1000 utterances of 8–10 seconds per language. This subset is used consistently in all experiments together with the AAL datasets.

\vspace{-5pt}
\subsection{Adaptation Tasks}
\vspace{-5pt}
We considered two adaptation scenarios for AALs. 

\noindent \textbf{Single AAL adaptation. }The goal is to distinguish a single AAL from HRLs. The multilingual LID model is adapted to the single target AAL while minimizing catastrophic forgetting, enabling it to capture language-specific acoustic features of the AAL while retaining correct identification of HRLs.

\noindent \textbf{Sequential adaptation. }The goal is to progressively adapt the pretrained LID model to multiple AALs while maintaining performance on HRLs and previously seen AALs. Each AAL is treated as an independent task and incorporated sequentially. For example, given three AALs $L_1, L_2, L_3$, the model is first adapted to $L_1$, and for each subsequent language $L_k$ ($k>1$), the model adapted to $L_{k-1}$ serves as the starting point. This approach enables cumulative learning, allowing incremental acquisition of multiple AALs while retaining prior AAL knowledge and discriminability between HRLs and AALs.

\vspace{-5pt}
\subsection{Implementation Details}
\vspace{-5pt}

\noindent \textbf{Pretrained model and adaptation strategy. } We adopt a pretrained model trained on the VoxLingua107 dataset \cite{valk2021voxlingua107}, which contains over 6,000 hours of speech from 107 languages, as the backbone for our experiments. The model comprises an ECAPA-TDNN (Emphasized Channel Attention, Propagation, and Aggregation Time Delay Neural Network) encoder \cite{ecapa} followed by a classifier\footnote{https://huggingface.co/speechbrain/lang-id-voxlingua107-ecapa} trained for 107-way language identification. Owing to its strong multilingual performance and robust architectural design, this model serves as a reliable baseline and is widely used in LID research.


During adaptation to new languages, the final classification layer of the pretrained model is replaced to include both newly introduced and previously learned languages in both proposed approaches. In RA-EWC, only the updated classifier is fine-tuned while the encoder remains frozen (Figure~\ref{fig:RA-EWC}). This strategy preserves prior acoustic representations and adjusting decision boundaries for new languages. In CG-KD, both the student model’s encoder and updated classifier are fine-tuned while the teacher model remains fixed (Figure~\ref{fig:2}), enabling representation-level adaptation guided by KD.

\noindent \textbf{Baselines. }We establish two types of baselines for comparison. First, conventional transfer learning, where pretrained models are fine-tuned on target low-resource language data without continual learning. Second, traditional CL approaches, including EWC \cite{EWC}, ER \cite{ER}, and KD \cite{KD}, with consistent data splits and model architectures across all experiments.

\noindent \textbf{Hyperparameters. }Each model is trained for 3 epochs with a batch size of 32 using the AdamW optimizer and a ReduceLROnPlateau scheduler (initial lr = 0.001). The hyperparameters $T$, $\alpha$, $\lambda$, and $\beta$ are tuned via grid search on the validation set and selected as $2.0$, $0.7$, $10^6$, and $1.0$, respectively. Performance is evaluated in terms of adaptation accuracy for AALs (F1-score on target AALs) and catastrophic forgetting (F1-score on 33 previously learned high-resource languages, \textit{HRL}). For the sequential adaptation task, forgetting is further assessed by including both HRL and the previously adapted AAL. Overall effectiveness is assessed jointly on the target AAL and HRL, referred to as \textit{overall}\footnote{https://github.com/PraviMyl/AAL\_identification}. 

\vspace{-10pt}
\section{Results and Discussion}
\label{results}

\subsection{Single AAL Identification}
\vspace{-5pt}
\begin{table*}[t]
\footnotesize
\centering
\caption{Performance (F1-score \%) of adapting proposed methods in Sequential using Warlpiri (WA), Dalabon (DA) and Dharawal (DH). Symbol $\cup$ denotes mixing training sets, while arrow $ \Rightarrow$ denotes training proposed methods sequentially. Best results in each method are highlighted in \textbf{Bold}.
}
\vspace{-5pt}
\begin{tabular}{lcccccc}
\hline
\textbf{Method} & \textbf{Settings} & \textbf{HRL (\%)} & \textbf{WA (\%)} & \textbf{DA (\%)} & \textbf{DH (\%)} & \textbf{overall (\%)} \\
\hline
\multirow{10}{*}{\centering RA-EWC} 
        &\textit{Baseline} & & & & &\\
       & WA $\cup$ DA $\cup$ DH    &  76.74 & 96.00 & 75.00 & 50.00 &  78.40 \\
       \cmidrule{2-7}
       &\textit{Sequential adaptation} & & & & &\\
       & WA $\Rightarrow$ DA $\Rightarrow$ DH & \textbf{77.86} & 94.57 & \textbf{100} & 100 & 79.12\\ 
       & WA $\Rightarrow$ DH $\Rightarrow$ DA &  77.57 & \textbf{98.26}      & 66.67 & 100 &  79.53 \\ 
       & DA $\Rightarrow$ WA $\Rightarrow$ DH &  77.91 & 91.16      & 75.00 & 100 &  80.13 \\ 
       & DA $\Rightarrow$ DH $\Rightarrow$ WA &  77.83 & 96.00      & 75.00 & 100 &  \textbf{80.20} \\ 
       & DH $\Rightarrow$ WA $\Rightarrow$ DA &  76.77 & 97.37      & 75.00 & 100 &  79.99 \\ 
       & DH $\Rightarrow$ DA $\Rightarrow$ WA &  75.23 & 96.46      & 75.00 & 100 &  78.69 \\

\midrule
\multirow{10}{*}{\centering CG-KD} 
       &\textit{Baseline} & & & & &\\
       & WA $\cup$ DA $\cup$ DH    &  87.70 & \textbf{98.70} & 82.35 & 66.67 &  87.50 \\
       \cmidrule{2-7}
       &\textit{Sequential adaptation} & & & & &\\
       & WA $\Rightarrow$ DA $\Rightarrow$ DH & 86.62 & 94.15 & \textbf{100} & 100 & 87.99\\ 
       & WA $\Rightarrow$ DH $\Rightarrow$ DA &  87.36 & 96.41      & 82.35 & 100 &  86.90 \\ 
       & DA $\Rightarrow$ WA $\Rightarrow$ DH &  88.18 & \textbf{98.70}      & 82.35 & 100 &  88.63 \\ 
       & DA $\Rightarrow$ DH $\Rightarrow$ WA &  87.61 & 91.82      & 82.35 & 100 &  86.76 \\ 
       & DH $\Rightarrow$ WA $\Rightarrow$ DA &  \textbf{89.06} & 96.12      & 82.35 & 100 &  \textbf{89.43} \\ 
       & DH $\Rightarrow$ DA $\Rightarrow$ WA &  86.13 & \textbf{98.70}      & 82.35 & 100 &  86.93 \\ 
\hline
\end{tabular}
\vspace{-10pt}
\label{tab:sequential}
\end{table*}

\noindent \textbf{LID performance for Warlpiri. }Table~\ref{tab:combined} shows the performance when Warlpiri is adapted to the pre-trained model. The pre-trained model (\textit{Source}) achieves an initial F1-score of 90.72\% on the high-resource languages (\textit{HRL}). A naïve transfer learning approach (\textit{TL}), where all model parameters are updated, performs well on Warlpiri but reduces the HRL F1-score to 62.47\%, indicating severe catastrophic forgetting.

Among the three baseline CL methods, KD performs best, achieving an overall F1-score of 93.02\% (91.56\% on HRL and 100\% on Warlpiri), effectively balancing knowledge retention and adaptation. EWC and ER also reduce forgetting, yielding more balanced results than TL. Our proposed RA-EWC further improves overall performance to 87.41\%, compared to 85.36\% and 85.26\% for EWC and ER, respectively, demonstrating improved stability–plasticity balance. The second proposed method, CG-KD, achieves the highest overall F1-score, increasing HRL performance from 91.56\% to 92.89\% while maintaining perfect adaptation to Warlpiri. This improvement is likely due to KD transferring richer information through soft targets, providing stronger regularization and more effectively mitigating catastrophic forgetting under low-resource conditions where replay buffers may be insufficient.


\noindent \textbf{LID performance for Dalabon and Dharawal: } Table \ref{tab:combined} further presents adaptation results for Dalabon and Dharawal, which are even more resource-constrained than Warlpiri. Both proposed methods effectively identify these extremely low-resource AALs while mitigating catastrophic forgetting. For Dalabon, EWC and ER both reduce forgetting compared to TL. KD achieves the best performance among the three baselines with an overall F1-score of 84.60\%, including 84.12\% on HRL and 100\% on Dalabon.  Our proposed CG-KD further improves the overall F1-score to 85.68\%, achieving perfect Dalabon adaptation while maintaining strong retention. A similar trend is observed for Dharawal, where both RA-EWC and CG-KD achieve perfect in-language classification (100\%) despite lower overall F1-scores (68.96\% and 76.41\%). These results demonstrate that the proposed methods can effectively adapt to extremely low-resource languages while maintaining meaningful HRL knowledge, highlighting the promise of CG-KD in highly data-scarce scenarios.

\vspace{-5pt}
\subsection{Sequential adaptation performance}
\vspace{-5pt}

Table~\ref{tab:sequential} presents F1-score of the proposed methods under various sequential adaptation scenarios, compared with a joint training baseline where all languages are learned simultaneously from a combined dataset. As described in Section~\ref{dataset}, the three AALs exhibit severe data imbalance, with Dharawal being extremely low-resource compared to Warlpiri and Dalabon.

Under the joint training baseline (\textit{WA $\cup$ DA $\cup$ DH}), both RA-EWC and CG-KD perform well on Warlpiri, but performance drops significantly for Dharawal, reaching only 50\% and 66.67\% F1-scores, respectively. This decline is mainly due to severe data imbalance, where joint optimization is dominated by majority language classes, leading to underperformance on extremely low-resource languages.

In contrast, sequential adaptation substantially improves performance for such languages. Treating Dharawal as an independent task allows the model to allocate dedicated capacity for learning discriminative, language-specific representations without interference from other languages. This approach does not depend on balanced datasets and enables the model to capture features that would otherwise be suppressed. Consequently, both RA-EWC and CG-KD achieve a perfect 100\% F1-score for Dharawal across all sequential scenarios. The performance gap between sequential adaptation and joint training highlights the effectiveness of the proposed approach in adapting to new languages under extremely low-resource conditions. 

Another key observation is the consistently high performance across all three AALs, regardless of adaptation order. For example, adapting Dharawal first, followed by Warlpiri and Dalabon, yields results comparable to other sequences. This indicates that the framework does not depend on dataset size, adaptation order, or closely related languages for effective learning, supporting flexible and long-term integration of new AALs as data becomes available.

Overall, CG-KD combined with sequential adaptation demonstrates strong robustness and scalability, providing a unified framework for identifying AALs among HRLs and multiple AALs under severe data imbalance. These results highlight its potential as a practical solution for building multilingual LID systems that can incrementally incorporate endangered AALs in real-world scenarios.
 

\vspace{-10pt}
\section{Conclusion}
\vspace{-5pt}

In this work, we propose two hybrid CL approaches, RA-EWC and CG-KD, to adapt pretrained speech foundation models for extremely low-resource AALs identification while mitigating catastrophic forgetting. Experimental results show that both methods outperform conventional fine-tuning and CL baselines, with CG-KD providing the strongest preservation of previously acquired knowledge and best performance under severe data scarcity. The proposed approaches also support sequential adaptation across multiple AALs, allowing models to evolve without degrading performance on earlier languages. Overall, these frameworks offer a scalable and robust solution for continual adaptation in severely under-resourced multilingual speech settings, advancing more inclusive language technologies.




\section{Acknowledgment}
The authors would like to thank the School of Electrical Engineering and Telecommunications at UNSW Sydney, Australia, for providing funding for this research initiative. Ethics approval was granted by the Human Research Ethics Advisory Panel Executive (HREAP) at UNSW Sydney, Australia, under Reference Number iRECS6867.

\section{Generative AI use disclosure}
During the preparation of this work, the authors used an AI tool in order to refine the academic language, improve the structural flow of the manuscript, and optimise technical terminology. After using this tool, the authors reviewed and edited the content as needed and take full responsibility for the content of the publication.

\bibliographystyle{IEEEtran}
\bibliography{refs}

\end{document}